\documentclass{article}
\usepackage{ijcai17}
\usepackage{times}

\usepackage{hyperref}
\usepackage{url}
\usepackage{amssymb}
\usepackage{amsmath}
\usepackage{graphicx}
\usepackage{chngcntr}
\usepackage[lined,algonl,boxed,oldcommands]{algorithm2e}
\usepackage{color}
\usepackage{subfigure}
\usepackage{lscape}
\usepackage[figuresleft]{rotating}
\usepackage{sidecap}
\usepackage[font=small]{caption}
\usepackage{amsthm}
\usepackage{booktabs, topcapt}
\usepackage{nicefrac}       
\usepackage{microtype}      
\usepackage{enumitem}

\newcommand{\be}{\begin{eqnarray} \begin{aligned}}
\newcommand{\ee}{\end{aligned} \end{eqnarray} }
\newcommand{\benn}{\begin{eqnarray*} \begin{aligned}}
\newcommand{\eenn}{\end{aligned} \end{eqnarray*} }
\newcommand{\la}{{\big \langle}}
\newcommand{\ra}{{\big \rangle}}
\usepackage{url}
\usepackage{color}

\renewenvironment{thebibliography}[1]{
  \begin{oldthebibliography}{#1}
    \setlength{\itemsep}{0em}
    \setlength{\parskip}{0em}
}
{
  \end{oldthebibliography}
}

\title{Unsupervised Learning via Total Correlation Explanation}

\author{
Greg Ver Steeg \\
University of Southern California \\
Information Sciences Institute \\
gregv@isi.edu
}

\begin{document}
\maketitle

\begin{abstract}
Learning by children and animals occurs effortlessly and largely without obvious supervision. 
Successes in automating supervised learning have not translated to the more ambiguous realm of unsupervised learning where goals and labels are not provided. 
Barlow (1961) suggested that the signal that brains leverage for unsupervised learning is dependence, or redundancy, in the sensory environment.
Dependence can be characterized using the information-theoretic multivariate mutual information measure called total correlation. 
The principle of Total Cor-relation Ex-planation (CorEx) is to learn representations of data that ``explain'' as much dependence in the data as possible.
We review some manifestations of this principle along with successes in unsupervised learning problems across diverse domains including human behavior, biology, and language. 
\end{abstract}

\section{Introduction} 

The brain is an information-processing chunk of meat with some amazing properties. Linsker, inventor of the InfoMax principle, made a statement thirty years ago that still rings true today, with some caveats.
\begin{quote}
A young animal or child perceives and identifies features in its environment in an apparently effortless way. No presently known algorithms even approach this flexible, general-purpose perceptual capability. Discovering the principles that may underlie perceptual processing is important both for neuroscience and for the development of synthetic perceptual systems.~\cite{linsker}
\end{quote}
Perception problems like visual object recognition that were once the exclusive domain of humans are now routinely carried out by computers. Surprisingly, this success did not come from profound understanding of the information processing principles in the brain, but more from brute force scaling of deep supervised optimization algorithms trained on large, labeled image datasets~\cite{bengioreview}.  On the other hand, supervised deep learners are not as flexible or general purpose as Linsker envisioned because they rely heavily on the availability and quality of the training labels. Moreover, the resulting representations can be brittle and hard to interpret~\cite{interpret}. Reducing the reliance on labels would greatly broaden the scope of these methods.

Unfortunately, successes in supervised learning have not translated to unsupervised learning~\cite{bengiounsupervised}. The inherent ambiguity and open-ended nature of the unsupervised problem makes it difficult to brute force a solution~\cite{minsky}. Learning principles that are general and flexible enough to apply across diverse domains, like human cognition, are needed. Barlow (1961) posited that the signal that the brain leverages for effective learning in the absence of direct supervision are redundancies or dependencies observed in the sensory environment. Investigating how the brain uses this redundancy has motivated several influential ideas~\cite{barlow_codes,simoncelli}.

This paper describes an information-theoretic approach to formalizing Barlow's idea called Total Cor-relation Ex-planation (CorEx). 
CorEx constructs a hierarchy of latent factors that progressively explain more dependencies in the observations as measured by multivariate information, also called total correlation. ``Explanation'' here is meant in the statistical sense that conditioned on the latent factors, $Y$, the observed random variables $X_1,\ldots, X_n$ will be statistically independent. 
While some learning approaches will learn to memorize even random, independent noise, from the CorEx perspective a lack of relationships in the data implies that there is nothing to learn. 
We will review the CorEx principle in the context of related information-theoretic methods then demonstrate its power and versatility on a wide range of unsupervised learning problems from human behavior to biology. 


\section{Information Principles for Learning}

Claude Shannon launched the field of information theory with a seminal paper in 1948. Consider a random variable, $X$, that can take values like $x$ with probability $p(X=x)$ (a coin flip, for example). Shannon defined the entropy of this random variable as $H(X) \equiv \la -\log p(x) \ra$ where brackets indicate expectation values over random variables. Shannon showed that this is the unique measure (up to scaling) that satisfies reasonable axioms and bounds the best rate of lossless compression. For a deeper intuition, see~\cite{dedeoit}. 
\nocite{shannon}

Given two random variables, $X_1$ and $X_2$, the mutual information is just the difference between the sum of individual entropies and the entropy of the variables considered jointly as a single system, 
$$I(X_1;X_2) \equiv H(X_1) + H(X_2) - H(X_1, X_2).$$ 
Shannon demonstrated that this quantity bounds the number of messages that can be reliably sent over a noisy channel.

The definition of mutual information between two parties can be generalized easily to $n$ parties. We refer to this generalization as $TC$ for total correlation following its historical introduction~\cite{watanabe}. 
\benn
TC(X_1, \ldots, X_n) &\equiv \sum_{i=1}^n H(X_i) - H(X_1, \ldots, X_n) \\
&= D_{KL} \left( p(x_1, \ldots, x_n) \| \prod_{i=1}^n p(x_i) \right).
\eenn 
In the following, we will often shorten $X \equiv X_1, \ldots, X_n$. We also wrote $TC$ in terms of KL-divergence, $D_{KL}$.  $TC(X)=0$ if and only if all the $X_i$ are independent. We can define conditional TC as $TC(X|Y) = D_{KL} ( p(x | y) \| \prod_i p(x_i|y) )$ and this quantity will be zero if and only if $X_i$ are independent conditioned on $Y$. 

Although Shannon gave a general yet precise definition of information, he warned that using everyday words for terminology can be misleading~\cite{bandwagon}.
If we consider building a representation of observations, $X$, as $Y=f(X)$, it seems quite intuitive to maximize the ``information'' that $Y$ has about $X$ by maximizing mutual information, $I(X;Y)$. Indeed, this is the substance of the InfoMax principle~\cite{linsker,bell95}. But the information is maximized if we simply memorize the data. Typically, InfoMax is invoked to maximize mutual information under some simplicity constraints. Still, it seems counter-intuitive that adding more resources should degrade learning performance. This phenomenon has been observed using InfoMax for clustering: more data leads to worse clustering~\cite{icml2014}. 

An opposite point of view is taken in the information bottleneck~\cite{tishby}. In that case, $I(X;Y)$ is actually minimized under the intuition that we should compress $X$ as much as possible. A trade-off is formulated that we should compress $Y$ while maintaining information about some labels, $Z$. While this approach is natural, it requires labeled data. 

Unsupervised approaches have also been motivated from the point of view that the brain has to compress information. In particular, Barlow points out that in a compressed representation neurons should fire independently~\cite{barlow_codes} (since correlations signal redundancy). In this spirit, independent component analysis (ICA) seeks to minimize $TC(Y)$, where $Y$ is a representation of the inputs, $X$~\cite{ica}. Other work suggests that efficient coding of information in the brain should additionally require that the firing of individual neurons be sparse~\cite{simoncelli}. While ICA has led to many useful results, the CorEx principle suggests a modification with several benefits. 

These learning principles are summarized in Table~\ref{tab:info}. While this list may look old-fashioned to some readers, these ideas are regularly invoked in modern papers on deep learning~\cite{bengio_autoencoders,nice,tishby_deep,NIB,infodropout,infogan}.
For brevity, other notable ideas have been omitted in this discussion including learning based on ``coarse-graining'' observations~\cite{licors,wolpert2014}, Jaynes' maximum entropy principle~\cite{jaynes}, integrated information theory~\cite{iit}, and common information~\cite{ijcai2017}.

\begin{table}[t]
\caption{Comparison of information principles for learning.}
\label{tab:info}
\vskip 0.05in
\begin{center}
\begin{footnotesize}
\begin{tabular}{ll}
\hline
Method & Objective \\
\hline
InfoMax  & $\max_{Y} I(Y; X)$ \\
Info bottleneck & $\min_{Y} I(Y; X) - \beta I(Y;Z)$ \\
ICA & $\min_{Y} TC(Y)$ \\
Common information & $\min_Y TC(X|Y)$ \\
CorEx (1 layer) & $\min_{Y} TC(X|Y) + TC(Y)$ \\
~~(1 layer, alternate form)& $\min_{Y} \sum_j I(Y_j;X) - \sum_i I(X_i;Y)$ \\
\hline
\end{tabular}
\end{footnotesize}
\end{center}
\vskip -0.1in
\end{table}

\subsection{Total Correlation Explanation}

One drawback of InfoMax, ICA, and others is that they are intrinsically shallow. For example, a transformation into independent components using ICA does not provide intermediate representations. Even if intermediate steps are used to get the independent components, there is no reason for these terms to be meaningful since they are not represented in the objective. 
It would be nice to have a hierarchy of abstraction where lower layers capture local relationships and higher layers reflect more global relationships. 

\begin{figure}[tbp] 
   \centering
   \includegraphics[width=0.95\columnwidth]{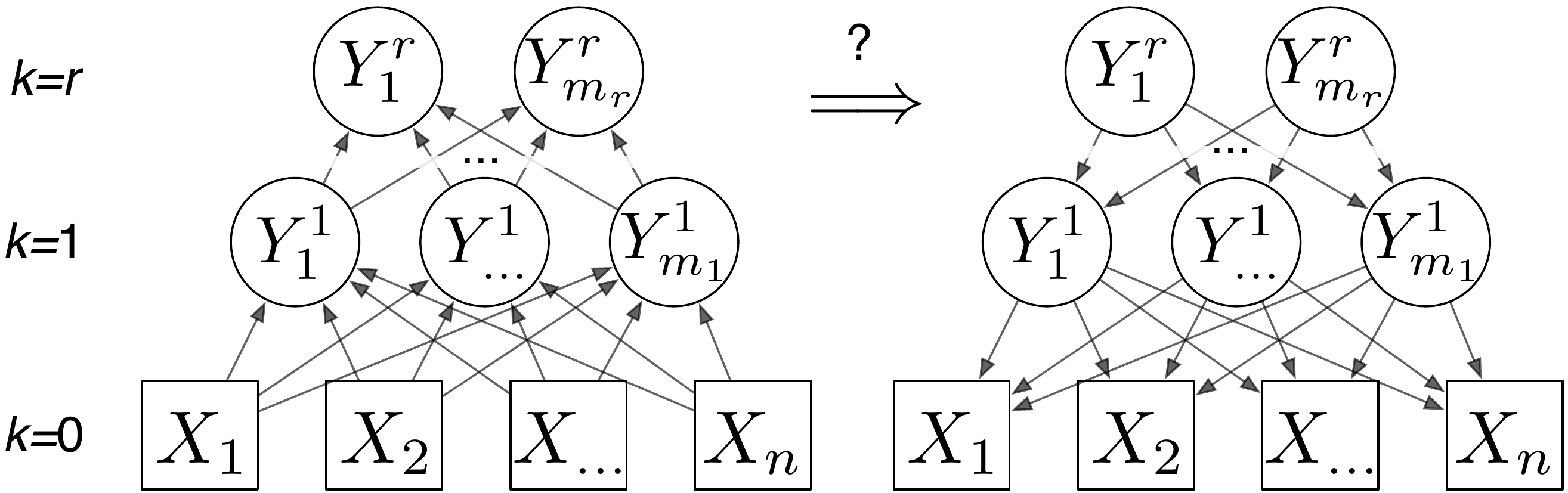} 
   \caption{A hierarchical representation where each layer is a probabilistic function of the layer below it. We deduce a condition under which we can flip the arrows and interpret the constructed factors, $Y$, as generating the dependence in the data.}
   \label{fig:hierarchy}
\end{figure}

\noindent\textbf{One Layer}\quad We will begin with a shallow formulation of CorEx, to see how it compares to other learning principles, then we will show that it has a natural hierarchical extension. 
Assume that each factor, $Y_j$, (also called neurons or hidden units) is a function of the data, $X$, $Y_j = f_j(X)$. More generally, it could be drawn from a probabilistic function, $Y_j \sim p(Y_j | X)$. The $Y$'s are generated according to a graphical model like the left side of Fig.~\ref{fig:hierarchy} (with $r=1$ for now). Under what conditions can we interpret these $Y$'s as latent factors that generate the data? In other words, when can we flip the arrows to get the graphical model on the right where the $Y$'s generate the dependence in $X$? The graphical model on the right side of Fig.~\ref{fig:hierarchy} is equivalent to a set of conditional independence relationships that can be summarized by saying $TC(X|Y) + TC(Y)=0$~\cite{nips2017}. In other words, each layer explains the correlations in the layer below or is independent. This is one way to write the CorEx objective (for one layer).
\benn
\min_{p(y_j|x)} TC(X|Y) + TC(Y)
\eenn
The objective is non-negative and the global minimum occurs at zero, in which case we can \emph{flip the arrows} to interpret $Y$'s as generating the dependence in $X$.  
Looking at Table~\ref{tab:info}, we see that this optimization has an ICA term plus another term that demands that $Y$'s make the $X$'s conditionally independent.
With some manipulation, it can be seen that this optimization is also equivalent to the following optimization~\cite{corex_theory}. 
\benn
 \min_{p(y_j|x)} \sum_j I(Y_j;X) - \sum_i I(X_i;Y)
\eenn
Again looking at Table~\ref{tab:info}, this alternate form shows the similarity to the information bottleneck. Instead of requiring labels, $Z$, we simply compress $X$ into each latent factor $Y_j$ while maintaining relevance about each of our observed variables, $X_i$. From this point of view, we can view CorEx as a special unsupervised version of the information bottleneck. 

\noindent\textbf{Multiple Layers}\quad Now we extend to the hierarchical case. We start by building a hierarchy with $r$ layers as on the left side of Fig.~\ref{fig:hierarchy}. For simplicity we define the variables at layer $k$ as $Y^k$ and we define $Y^0 \equiv X$ and $Y^{r+1} = 0$ (a constant).  Now let each layer explain dependence in the layer below. 
\be\label{eq:opt1}
\min_{p(y_j^k|x)} \sum_{k=0}^r TC(Y^{k}|Y^{k+1}) 
\ee
Again, this quantity is non-negative and has a global minimum at zero. At the global minimum, we can flip the arrows and interpret the latent factors as a generative model for the dependence in $X$.
We can again re-write the objective as a sum of bottleneck-like optimizations at each layer of the hierarchy.
\be
 \max_{p(y_j^k|x)} \sum_{k=0}^r \left(\sum_i I(Y_i^k;Y^{k+1}) - \sum_j I(Y_j^{k+1};Y^k)  \right)
\ee
An advantage of writing the objective in this second form is that it can be directly plugged into a useful inequality~\cite{corex_theory}. 
\be\label{eq:bound}
TC(X) \geq \sum_{k=0}^r \left(\sum_i I(Y_i^k;Y^{k+1}) - \sum_j I(Y_j^{k+1};Y^k)  \right)
\ee
$TC(X)$ is the amount of dependence observed in the data. For high-dimensional systems, this is hard to estimate. However, the bound in Eq.~\ref{eq:bound} allows us to solve a hierarchy of optimization problems giving progressively tighter lower bounds on the dependence in $X$. Because each layer directly contributes to the lower bound on $TC(X)$, we can quantify the value of depth and stop adding layers to the hierarchy when the lower bound stops increasing. Local optima of this non-convex optimization can be found using algorithms with low computational and sample complexity.

CorEx hierarchically decomposes multivariate information in $X$ in terms of contributions from latent factors at each layer of a hierarchy. This decomposition can be viewed as a generalization of the hierarchical decomposition introduced by Watanabe~\cite{watanabe}. 

\noindent\textbf{Incremental CorEx}\quad Just as PCA has an incremental version where one component is extracted at a time, the incremental version of the CorEx principle is the information sieve~\cite{sieve}. For the incremental version, we consider the special case where $Y$ is one-dimensional so that $TC(Y)=0$ and the objective in Eq.~\ref{eq:opt1} reduces to $\min_Y TC(X|Y)$. After learning one $Y$ that optimizes the objective, we transform the data (a kind of information-theoretic orthogonalization) so that we can learn another factor that extracts more dependence. 
The sieve optimization is a dual formulation of the optimization defining the Wyner common information~\cite{wyner_common,gastpar} and can be viewed as a decomposition of common information~\cite{ijcai2017}. 


\subsection{Implementations of the CorEx Principle}

We briefly review implementations of the CorEx principle, including their applicability and limitations. 
Code is available at \url{http://github.com/gregversteeg}. 
\begin{itemize}[noitemsep,topsep=0pt,parsep=0pt,partopsep=0pt]
\item \url{CorEx}~\cite{nips2014}: The original implementation is restricted to discrete variables and tree structured latent factors. The functionality is subsumed by other versions. 
\item \url{bio_corex}~\cite{corex_theory,pepke}: The most flexible version includes augmentations that were designed for challenges in the biology domain, though there is nothing specific to biology in the implementation. This version handles discrete and continuous variables, overlapping latent factor structure, and missing data.  Although CorEx has intrinsically low sample complexity, some biology data is severely under-sampled. This version implements a Bayesian smoothing of the marginal parameter estimates that reduces the appearance of spurious correlations~\cite{pepke}. This version runs quickly for problems with up to thousands of variables.
\item \url{discrete_sieve}~\cite{sieve}: This is the first implementation of the information sieve for discrete variables. In the discrete case, the information orthogonalization required by the sieve is extremely challenging. The approach is impractical for most real world problems. 
\item \url{LinearSieve}~\cite{ijcai2017}:  The linear version of the information sieve (for continuous variables) is fast and practical for finding the top components explaining the most correlation in data. Like other incremental methods, repeated application eventually introduces numerical instability. 
\item \url{LinearCorEx}~\cite{nips2017}:  The fastest, non-sparse version of CorEx assumes latent factors are linear functions of the inputs. Linear CorEx is very effective for covariance estimation and subspace clustering in high-dimensional, under-sampled data. 
\item \url{corex_topic}~\cite{reing2016,gallagher}: This implementation exploits sparsity for major speed-ups, easily handling hundreds of thousands of variables depending on the sparsity of the data. While this version was developed for topic modeling using binary bag of words data, it can be applied to any binary data. This implementation includes a semi-supervised option. 
\end{itemize}
\section{Applications}
The most successful applications of the CorEx principle involve high-dimensional data in complex domains that are difficult to model a priori. In practice, the learned hierarchical representation is used in several ways. First, the structure induces a hierarchical clustering of input variables. Second, the latent factors at each layer constitute a reduced dimensionality representation of the input data. Third, individual factors often disentangle true latent factors of variation in the data. Finally, like PCA, we can rank factors according to their value. While PCA finds components that explain the most \emph{variance}, in CorEx we find factors that explain the most \emph{dependence}. 

\noindent\textbf{Social Science}\quad 
Latent factor models like the ``Big 5'' personality factors are popular in social science because of their simplicity and interpretability. Given data from an online survey with the Big 5 questions, CorEx was able to perfectly reconstruct the Big 5 factors while standard methods failed~\cite{nips2014}. CorEx outperforms traditional factor methods when the number of variables is larger than the number of samples~\cite{ijcai2017}, as is increasingly common in social science experiments. 

\noindent\textbf{Gene Expression}\quad  
If DNA is the software for our biology, then gene expression tells us which code is running. New sequencing methods allow us to read expression levels for thousands of genes but, unfortunately, only a tiny fraction of these processes are understood. Applying CorEx to gene expression reveals a rich array of strong biological signals. Gene clusters discovered by CorEx exhibit four times more enrichment with respect to known biology in gene ontology databases than standard approaches~\cite{pepke}. The CorEx hierarchy also seems to accurately reflect biological organization. For instance, two low-level clusters, immune cell activation and inflammatory response, are combined in a higher-level group related to immune signaling. 

\noindent\textbf{Neuroscience}\quad 
Next generation imaging technology gives detailed views of individual brains but even the largest studies can only afford to image a relatively small number of brains. Small sample sizes with high-dimensional data has led to low statistical power in most studies and a crisis of replication in the field~\cite{powerfailure}. 
Exploratory data analysis using CorEx can help identify only the strongest, most robust dependencies in the data, even with small sample sizes. Early results have identified well-known relationships along with novel, biologically plausible candidate effects that increase predictive power~\cite{isbi_blood,isbi_value,artemis}. Moreover, CorEx outperforms ICA for disentangling spatial modes in imaging data~\cite{ijcai2017}.

\noindent\textbf{Language} 
Applied to bag of words vectors of text data, CorEx representations can be interpreted as hierarchical topic models~\cite{nips2014,hodas2015}. 
Latent tree based methods like CorEx outperform LDA~\cite{chen_topic} on several measures of topic quality. A semi-supervised version further increases the interpretability of topics by allowing us to ``anchor'' some latent factors to designated words of interest~\cite{reing2016,gallagher}.

\noindent\textbf{Finance}\quad  
The stock market exhibits a high degree of dependence and measuring this dependence is important for quantifying risk. Applied to monthly returns on the S\&P 500, CorEx discovers a hierarchy of latent factors related to industry sectors~\cite{corex_theory}. Using a linear version of CorEx allows us to estimate covariance matrices and this approach outperforms state-of-the-art methods like GLASSO for under-sampled, high-dimensional stock market data~\cite{nips2017}.

\section{Conclusion} 

Learning useful representations of observations without supervision across diverse domains is a challenging, unsolved problem. 
While successes in supervised learning have not immediately translated to comparable results for unsupervised learning, the engineering achievements that have driven those successes now allow us to define flexible, powerful architectures and quickly and easily optimize them under a wide variety objectives~\cite{tensorflow_short}. 
Taking information-theoretic learning principles and applying them within these powerful frameworks has already generated many compelling results~\cite{infogan,NIB,infodropout}. Applying the CorEx principle in the same context is a logical step for future work. 

Unlike similar learning principles, CorEx naturally decomposes information in a hierarchical way. This hierarchical structure has numerical advantages since the objective is less reliant on back-propagation for training. Credit for predicting a correct label does not have to be assigned to intermediate representations as in supervised learning~\cite{minsky}, instead each latent factor in the representation has a quantifiable information value within the hierarchical decomposition. Besides the technical advantages, many applications show that individual latent factors learned via CorEx reflect diverse and meaningful structure in real-world datasets. We hope that ongoing CorEx developments continue to accelerate the discovery of knowledge through unsupervised learning. 


{\small
 \bibliographystyle{named}
\bibliography{gversteeg_bibdesk} 
}

\end{document}